\documentclass{article}

\usepackage{arxiv}

\usepackage[utf8]{inputenc} % allow utf-8 input
\usepackage[T1]{fontenc}    % use 8-bit T1 fonts
\usepackage{hyperref}       % hyperlinks
\usepackage{url}            % simple URL typesetting
\usepackage{booktabs}       % professional-quality tables
\usepackage{amsfonts}       % blackboard math symbols
\usepackage{nicefrac}       % compact symbols for 1/2, etc.
\usepackage{microtype}      % microtypography
\usepackage{lipsum}
\usepackage{graphicx}
\usepackage{caption}
\usepackage{subcaption}
\usepackage{multirow}
\usepackage{hyperref}

\title{ML Health: Fitness Tracking for Production Models}

\author{
  Sindhu Ghanta\\
  ParallelM\\
  \texttt{sindhu.ghanta@parallelm.com} \\
   \And
 Sriram Subramanian \\
  ParallelM\\
  \texttt{sriram.subramanian@parallelm.com} \\
  \And
 Lior Khermosh \\
  ParallelM\\
  \texttt{lior.khermosh@parallelm.com} \\
  \And
 Swaminathan Sundararaman \\
  ParallelM\\
  \texttt{swami@parallelm.com} \\
  \And
 Harshil Shah \\
  ParallelM\\
  \texttt{harshil.shah@parallelm.com} \\
  \And
 Yakov Goldberg \\
  ParallelM\\
  \texttt{yakov.goldberg@parallelm.com} \\
  \And
  Drew Roselli \\
  ParallelM\\
  \texttt{drew.roselli@parallelm.com} \\
  \And
  Nisha Talagala \\
  ParallelM\\
  \texttt{nisha.talagala@parallelm.com} \\
}

\begin{document}
\maketitle

\begin{abstract}
Deployment of machine learning (ML) algorithms in production for extended periods of time has uncovered new challenges such as monitoring and management of real-time prediction quality of a model in the absence of labels. However, such tracking is imperative to prevent catastrophic business outcomes resulting from incorrect predictions. The scale of these deployments makes manual monitoring prohibitive, making automated techniques to track and raise alerts imperative. We present a framework, {\it ML Health}, for tracking potential drops in the predictive performance of ML models in the {\it absence of labels}. The framework employs diagnostic methods to generate alerts for further investigation. We develop one such method to monitor potential problems when production data patterns do not match training data distributions. We demonstrate that our method performs better than standard "distance metrics", such as RMSE, KL-Divergence, and Wasserstein at detecting issues with mismatched data sets. 
Finally, we present a working system that incorporates the ML Health approach to monitor and manage ML deployments within a realistic full production ML lifecycle. 
\end{abstract}

% keywords can be removed

\section{Introduction}
Machine Learning (ML) models are being increasingly deployed in production in a range of domains spanning advertising, recommendation engines, medical prognosis etc. The motivation of adapting machine learning in any business is to enable automation and intelligent decision making, typically driven by learnings from historical data. 

The typical lifecycle of deployment of these models involves a training phase, where a data scientist develops a model with good predictive performance, based on historical data. This model is put into production with the hope that it would continue to have similar predictive performance during the course of its deployment. The above process is problematic in production where the following scenarios could happen: (a) an incorrect model gets pushed (b) incoming data is corrupted (c) incoming data changes and no longer resembles datasets used during training \cite{modelSelection,driftEnsemble}. In all of these cases, traditional software methods for detecting failures will not be effective. The production code will continue to run and may generate sub-optimal predictions with no visible indicator of fault. 

As ML algorithms permeate daily life, scenarios of poor predictions causing damage are increasingly reported. Examples include financial loss ~\cite{knight-capital}, loss of human health and life ~\cite{michael_jordan}, and corporate embarrassment due to allegations of bias ~\cite{dastin, vincent}. Industries and governments are also imposing increasing regulations, requiring businesses to manage the quality of their product's ML predictions ~\cite{newyork2018, gdpr1_2018, sr_11_7}. 

While the issues described above vary, they all point to a single core problem which is that production environments can change dramatically relative to the more pristine training environment. When these changes occur, ML models that scored well in training exercises can behave in unpredictable and sometimes harmful ways.

To further illustrate this issue, we describe an experiment where a well-established and classic ML algorithm (Random Forest) is used to predict SLA violations from a publicly available TELCO dataset \cite{telco}. It inherently consists of different types of loads (Periodic, Flash and Linear) that might be seen in production. Each load has the same features but differs in feature distribution which has varying impact on the performance of a model that is trained on a different type of load. Table \ref{tab:telco} shows how the algorithm, trained on different workload patterns, behaves when shown other patterns that were not part of its training experience. As the table shows, when the algorithm is presented with data similar to what it saw during training, predictive performance is good. When the incoming production dataset varies, the predictive performance also varies in different ways depending on how the incoming data differed from the training set. Oftentimes, these changes in incoming data are not foreseeable and hence cannot be accounted for during the training phase. 

We present a framework {\it ML Health} that can be used to detect the above scenarios and generate alerts for further investigation. Our work makes the following contributions:

\begin{itemize}
\item We motivate the need for ML Health metrics and indicators in production ML workflows
\item We describe a new ML Health indicator, Similarity, and demonstrate how it outperforms known state of the art.
\item We describe a practical system implementation of this indicator and experiences with production use. 
\end{itemize}

\if 0 
Alerts will then notify operations personnel who can bring data scientists in for further investigation. In addition, a system that supports feedback from production deployment is necessary to enable quick debugging and smooth operation of the ML pipeline over extended periods of time.
\fi

% DREW: experiment is not "artificial"
% DREW: Table ??

\if 0
%% comment for Sindhu - one thing I do when using the Telco illustration is to have two versions of the SVM algorithm. I use this to point out not just that the algorithm behaves poorly when show a dataset unlike its training - but even the degree to which is degrades is unpredictable. The two SVM variants do not even degrade i the same way. This then drives the conclusion that the solution is not to fix the algorithm - its to detect the data deviation in the first place. If you can add that to the table - we can add a paragraph here pointing that out
\fi

\begin{center}
    \begin{table}
    \centering
    \label{tab:telco_accuracy}
    \begin{tabular}{ |l | l | l | l| }
    \hline
    Load Type & Periodic & Flash & Linear  \\ \hline
    Periodic &  0.87 & 0.6 & 0.36  \\ \hline
    Flash & 0.35 & 0.91 & 0.81 \\ \hline
    Linear & 0.35 & 0.53 & 0.86  \\ \hline
    \end{tabular}
    \caption{Accuracy values for different combination of loads. Each row corresponds to the training dataset. Each column corresponds to the inference dataset. When incoming data looks like training data, the algorithm does well. When the incoming data pattern deviates from training data, the algorithm struggles and predictive performance varies.\label{tab:telco}}
    \end{table}
\end{center}

An attempt at solving the above problem has been made mainly in the streaming scenarios. There exists extensive literature on the concept of covariate/concept shift/drift \cite{modelSelection,driftEnsemble}, where the training and inference datasets have different characteristics leading to poorly performing machine learning models. When the features of the dataset drift, its called covariate shift, while a shift in the relationship between the features and the target variable is called concept shift \cite{learningTransfer}. Note that detection of concept shift requires the presence of labels. Typically, these methods also assume labels are available shortly after making the prediction and hence the classifier error rate can be monitored. They either train continuously \cite{passive} from new data or monitor classifier error rate continuously \cite{EDDM}. \cite{drift-unlabelled} tracks drift by detecting the number of samples that are in the uncertainty region of the classifier, assuming it is possible to obtain this boundary. Our research differs in a number of ways
(a) we assume that no labels are available. This makes our approach more generally usable in production scenarios, many of which cannot generate labels in real time
% DREW b) is not parseable
(b) our approach is algorithm agnotsic (demonstrated later for classification and regression), 
(c) our technique  can be applied to streaming as well as batch scenarios. However, our technique does not focus on detecting evolving/changing relationship between data and labels unlike the drift detection techniques. 

We propose to monitor the change in patterns of incoming features in comparison to the ones observed during training and argue that such a change could indicate the fitness of a ML model to inference data. This eliminates the need for availability of labels. There exist techniques such as KL-divergence \cite{kullback1951}, Wasserstein metric \cite{wasserstein}, etc. that provide a score for divergence between two distributions. However, they rely on the fact that the inference distribution is available and representative of the inference data. This implies there are enough samples to form a representative distribution \cite{multi-sample}. 

To overcome this limitation, we  present an approach that is agnostic of the number of incoming samples. We elaborate on the details of our technique in Section \ref{subsec:dd}. Finally, as part of a robust approach to ML Health, we provide an interface and library for users to create their own statistics and response functions. Such functions can be used to generate alerts to drive  corrective action, such as model rollback. %%Needs proper connection

% DREW: with simple in-built metrics
Finally, we demonstrate our ML health metric in the context of a complete production ML solution. Our system design supports large-scale, distributed machine learning production deployments while providing the flexibility for application and algorithm specific customization.

\section{ML Health}
\label{sec:health}
We consider a system to have good {\it ML Health} if the deployed model has good predictive performance.
In this section, we present an {\it ML Health} approach called {\it{Similarity}} score that detects a deviation in the incoming features compared to training.

\subsection{Motivation}
\begin{figure}[htp]
    \centering
    \begin{subfigure}{0.42\textwidth}
        \centering
        \includegraphics[trim={0 0 4cm 0},height=4cm,clip]{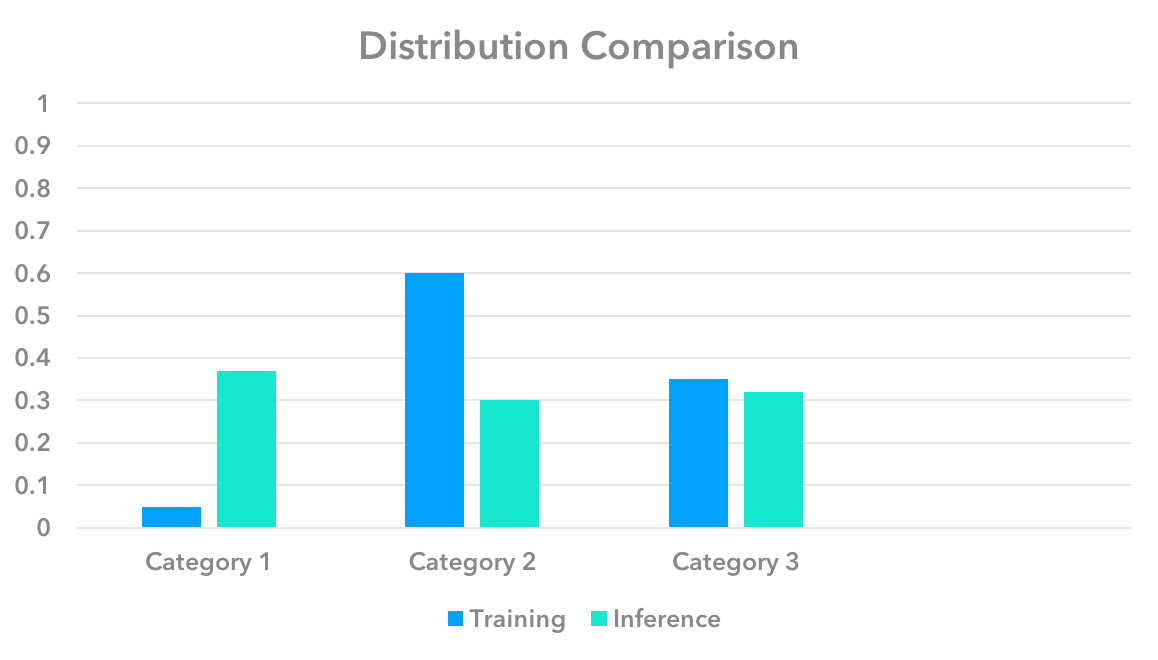}
        \caption{KL-divergence (0.34), RMSE (0.41), Similarity (0.64)}
        \label{fig:case1}
    \end{subfigure}
    \begin{subfigure}{0.42\textwidth}
        \centering
        \includegraphics[trim={0 0 4cm 0},height=4cm,clip]{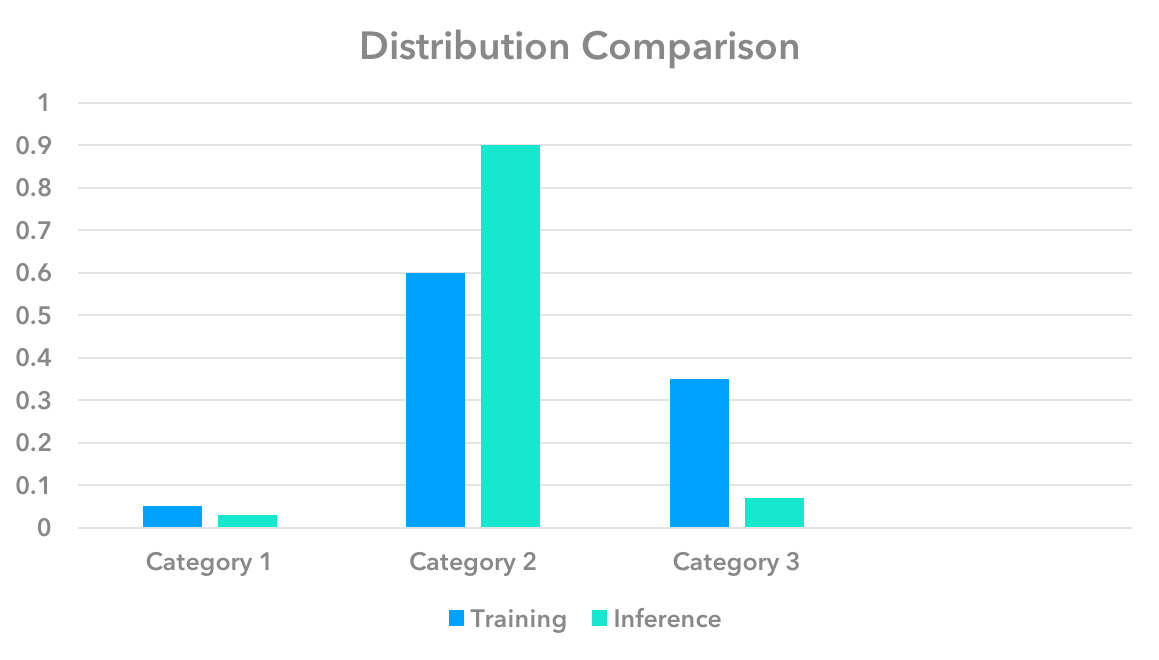}
        \caption{KL-divergence (0.34), RMSE (0.43), Similarity (1.0)}
        \label{fig:case2}
    \end{subfigure}
    \caption{Training vs Inference distributions}
    \label{fig:cases}
\end{figure}

Popular techniques used for calculating the similarity between two distributions are KL-divergence, RMSE, etc. These techniques assume the availability of two representative distributions and present different ways to quantify divergence between them. However, there are three issues with this approach (a) the number of samples available for inference can vary widely, making the creation of representative inference distributions hard (b) it is possible that all the inference samples fall in a narrow range of the distribution seen during training. It indicates that only a subset of the patterns seen during training are occurring in inference. Such a scenario should not be penalized since it does not indicate a divergence from the patterns seen during training. (c) The subset of patterns seen during inference might either have a poor training coverage or a good training coverage. We demonstrate case (c) using a concrete example. 
%% DREW: gets confusing which frequently goes with which training/inf
%% I suggest combining next 2 paragraphs and cover each case
%% completely. Another suggestion is to name these concepts:
%% poor training coverage (not well represented in training
%% good training coverage (well represented in training
Figure \ref{fig:case1} and \ref{fig:case2} show normalized training versus inference distributions in two cases for $1$ feature. Let this feature contain $3$ categories. Note that the training distribution in both figures is identical. In the training distribution/histogram, it can be observed that Category 1 is under-represented, while Category 2 has a relatively high occurrence in training. 

The inference distribution/histograms in the two figures represent two types of scenarios (a) Poor training coverage: Relative frequency of Category 1 in training is small compared to other categories. However, during inference, Category 1 shows a much higher occurrence (Figure \ref{fig:case1}) and (b) Good training coverage: Category 2 has high occurrence in training compared to other categories. During inference, Category 2 occurs at a much higher frequency compared to training (Figure \ref{fig:case2}).

 Typically, the Poor training coverage scenario described in Figure \ref{fig:case1} is more concerning than Good training coverage scenario shown in Figure \ref{fig:case2}, especially when such a scenario is extended to 1000s of categories. The intuition behind this is that the algorithm has a chance to train well on a category that is relatively well represented during training (which is Category 2 in this example). A high occurrence of Category 2 is therefore not of great concern. However, the algorithm did not see Category 1 very often during training. When this category shows up much more often, it is of concern as one cannot be sure of the robustness of the algorithm's predictive performance for such a poorly represented pattern.

 Therefore, a metric that alerts on the poor training coverage case, rather than good training coverage case is of greater interest. Standard distribution comparison metrics such as KL-divergence and root mean square error (RMSE) are not designed to capture such an asymmetric deviation, as can be seen in the example illustrated in Figure \ref{fig:cases}. To mitigate the issues described above, we propose the {\it Similarity} score that has been described in detail in the subsection {\bf Similarity Score}.
  %% Drew: fill in section 

\subsection{Similarity Score}
\label{subsec:dd}
In this section, we describe the calculation of the {\it Similarity} score that indicates the likelihood of a sample to belong to a given distribution. 

Let ${\bf X_T} \in \mathbb{R}^{N_T \times M}$ represent the training dataset, where $N_T$ is the number of samples and $M$ is the number of features. Let $x_{T,m}$ represent the $m^{th}$ feature/column of the training dataset. Let each feature be represented by an independent multinoulli distribution. In case of continuous features, they are discretized using bins and then modelled as a multinoulli distribution. Let the number of categories in each feature be given by $[C_1,C_2,...,C_M]$, where $M$ is the number of features. The probability of occurrence of any particular category value for a feature is given by $p(x_{T,m}) = \sum_{c=1}^{C_m} p_{c_m}^{[x_{T,m}=c]}$, where $p_{c_m}$ is the probability of $m^{th}$ feature to belong to category $c$. 
%% DREW: this is hard to parse. Break into another sentence
$[x=c]$ evaluates to $1$ if $x=c$, $0$ otherwise. We drop $m$ from notation for simplicity.

Let the frequency of occurrence of each category in a single feature in training be given by $[f_{T,1},f_{T,2},..,f_{T,C}]$, where C is the total number of categories in this feature and $\sum_{i=1}^{C} f_{T,i} = N_T $. Therefore, $p_c = f_{T,c}/N_T$ is the parameter of the multinoulli distribution. Let ${\bf X_I} \in \mathbb{R}^{N_I \times M}$ represent the inference dataset, where $N_I$ is the number of samples and $M$ is the number of features. Let the frequency of occurrence of each category in a single feature (in inference) be given by $[f_{I,1},f_{I,2},..,f_{I,C}]$, where C is the total number of categories in this feature and $\sum_{i=1}^{C} f_{I,i} = N_I $. The average probability across $N_I$ samples for a single feature is given by:
\begin{equation}
\label{eq:inference}
p(X_I) = \frac{1}{N_I}\sum_{i=1}^{N_I} p(x_{i})\\
= \frac{1}{N_T\times N_I} \sum_{i=1}^{C}f_{T,i}\times f_{I,i}
\end{equation}

Equation \ref{eq:inference} could run into numerical issues especially in the presence of a large number of categories. To mitigate this issue, we add a normalization factor. We choose this factor to be the average training probability across $N_T$ samples since this gives us a score that is relative to the distribution observed during training. As a result of this normalization, it is possible to get a $p(X_I)$ value that is greater than one. Such a score indicates that the frequently occurring categories have occurred much more in inference relative to training.

\begin{equation}
\label{eq:training}
p(X_T) = \frac{1}{N_T^2} \sum_{i=1}^{C}f_{T,i}^2
\end{equation}

The normalization factor is given by Equation \ref{eq:training}. Equation \ref{eq:norm} is the expression obtained after adding the normalization factor to Equation  \ref{eq:inference}. This similarity score can be used to detect divergence in distributions. 
\begin{equation}
\label{eq:norm}
p(X_I) = \frac{N_T}{N_I} \times \frac{\sum_{i=1}^{C}f_{T,i}\times f_{I,i}}{\sum_{j=1}^{C}f_{T,j}^2}
\end{equation}

The score presented in this section addresses the three issues pointed out in the {\bf Motivation} subsection. (a) low number of samples: since the {\it Similarity} score is calculated based on the parameters of the multinoulli distribution and does not rely on the inference distribution, it is agnostic of the number of samples (b) inference samples fall in a narrow range of training distribution: {\it Similarity} score reply on the probability values associated with this narrow range of distribution and hence does not penalize the fact that inference distribution does not cover the entire range of categories observed during training (c) The subset of patterns seen during inference might either have a poor training coverage or a good training coverage: {\it Similarity} score is designed to tackle this scenario using the training probability values associated with each category. Similarity score reported in Figure \ref{fig:case1} vs \ref{fig:case2} demonstrates that the score penalized a poor training coverage scenario and not the good training coverage scenario.

In addition to a score for detecting potential drops in predictive performance, the system infrastructure must support such a feedback and alerting mechanism and ideally handle the diversity of engines and languages typically used for ML applications (Spark, python, R, etc.). In Section \ref{sec:sysOverview}, we describe a system that leverages this score to generate alerts in production deployments.

\section{System Overview}
\label{sec:sysOverview}
We have built MCenter, a distributed system to help data scientists build and deploy ML/DL applications in production. We enable reporting of custom metrics such as the Similarity score by providing an API (called MLOps, described in Section \ref{sec:mlops}) along with a centralized location for visualization and monitoring that is agnostic of the engine/language. MCenter supports a gamut of analytic engines/libraries (such as Spark, TensorFlow, Scikit-Learn, and PyTorch), languages (such as R, Python, Scala, and Java), resource managers (such as YARN, Kubernetes, Docker Server, Stand-alone Spark, and Flink) and Cloud deployments (such as Azure, AWS, and GCP). MCenter has been in production for over a year and has been helping numerous data scientists to deploy and manage ML/DL based initiatives with confidence.

\subsection{Architecture}

\begin{figure}[http!]
\centering
\includegraphics[trim={0cm 0.1cm 0cm 0cm},clip,width=0.42\textwidth]{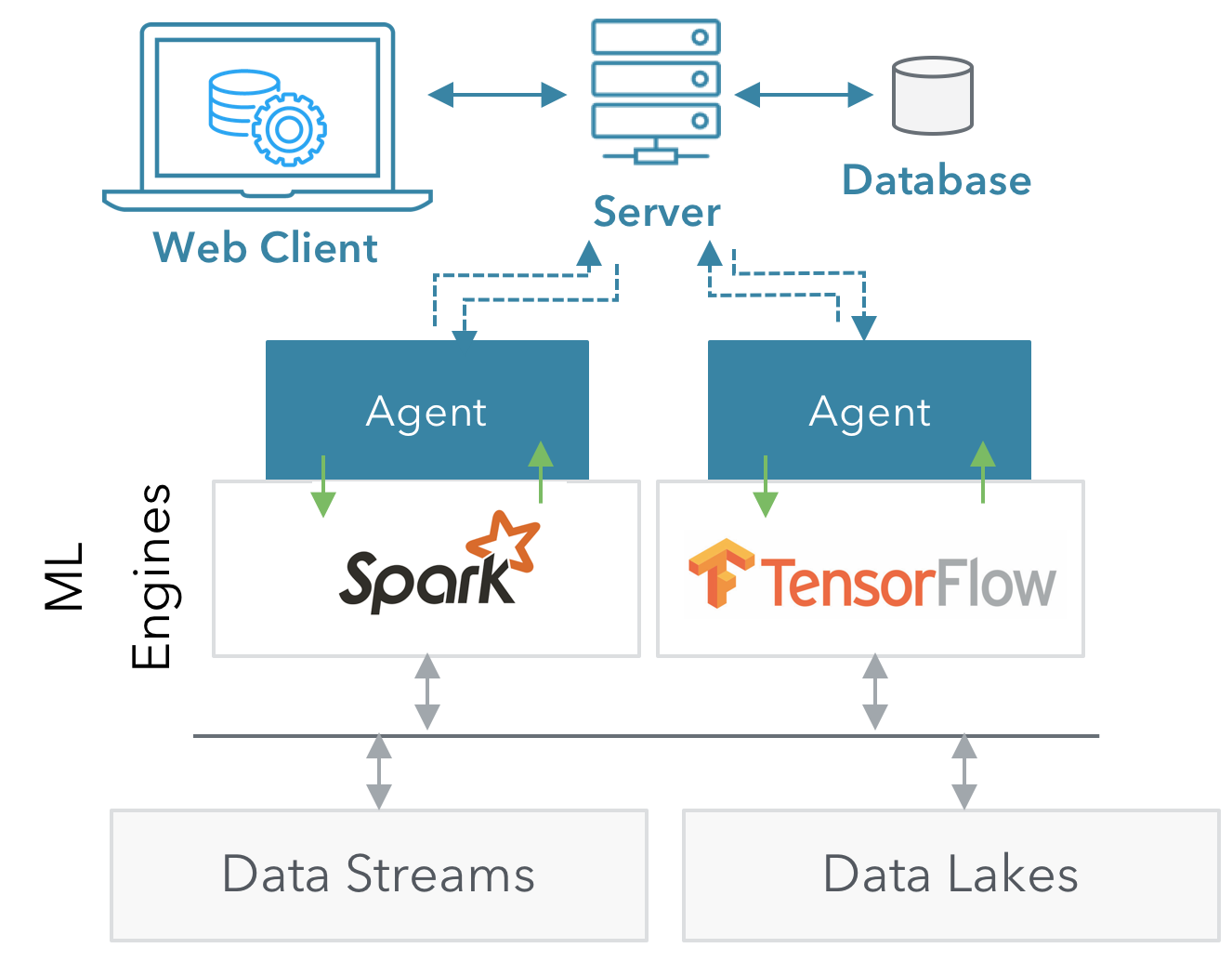}              
\caption{Architecture of our Solution \label{fig:Arch}}
\end{figure}

A high-level architecture overview is provided in Figure~\ref{fig:Arch}. At the core of this architecture is a Server and an Agent. The Server keeps track of all the deployments across agents, while the agents themselves are agnostic of all the jobs running in the system. Details of this architecture can be found in \cite{Governance2018}. 

Users can leverage built-in applications and health metrics, create new applications and/or metrics, or import their existing ML pipelines and health-metric computations using any of the popular programming languages. They can use the MLOps API to report custom statistics which are stored in the central database. During inference, along with the model, all associated statistics can be retrieved and displayed in a central location. Typically, in the absence of such a system for production, the data scientist will need to get the system logs from an operations engineer and filter through them to get the relevant information.

\begin{table*}
\begin{center}
\label{tab:mlstats-api}
    \begin{tabular}{ | c | p{4cm} | p{10cm}| }
    \hline
    \bf{Category} & \bf{Usage} & \bf{API} \\ \hline
    {\em{Metrics/Statistics}} & Store/set and Fetch/get metrics and statistics & set\_stat(name, data=None, category), {\it {set\_data\_distribution\_stat(data, model=None)}}  \\ \hline
    {\em{Models}} & Store and fetch models & current\_model() get\_models\_by\_time(start\_time, end\_time\\ \hline
    {\em{Notification}} & Help notify results / status & health\_alert(title, description, data=None)\\
    \hline
    \end{tabular}
\caption{Simplified version of the API to help import and export objects from the DB}
\label{tab:api}
\end{center}
\vspace{-5mm}
\end{table*}

\subsection{MLOps API Framework}
\label{sec:mlops}
To support heterogeneity (i.e., programming languages and analytic engines) and flexibility to write custom health metrics using our system, we provide an MLOps library that has the ability to gather information from each executing pipeline (see Table \ref{tab:api}). Via this API, ML pipelines can report a variety of objects and statistics, from models to accuracy metrics, alerts, and event notifications. 

The ML Health techniques presented in this paper use this instrumented information to perform ML Health assessments. For example, the histograms calculated during the training part of an ML application are reported into the database using the {\em export} part of the MLstat API. These statistics are fetched by the inference pipeline using the {\em import} part of the MLstat API. Using this information, any metric can be calculated from the incoming data and reported back into the database using the same API. Similarity score is provided as a default health metric that is calculated and stored in the system using the 
{\it {set\_data\_distribution\_stat}} API. Alerts can be configured based on this similarity score and reported using the MLOps API. Complex logic can also be embedded into how these alerts should be generated based on any unique use case requirements.

A complete description of all the APIs is provided in the documentation of the MCenter software that is available for free download at \hyperref[]{``https://www.parallelm.com/free-account/''}

\section{Experiments}
We compare our {\it Similarity} score with established distribution comparison techniques (KL Divergence, Wasserstein Metric) and the raw probability value reported by many classification algorithms. In Section {\bf Dataset Description}, we describe the datasets used for the experiments followed by Section {\bf Experimental Setup} that describes the steps of the experiment in detail. We report the results of these experiments in Section {\bf Results} followed by a description of deployment in MCenter system in Section {\bf System Deployment and Operations}.

\subsection{Dataset Description}
\label{subsec:datasets}
We show results on three publicly available datasets (a) Human Activity Recognition (Samsung) \cite{Samsung} (b) Video Transcoding Time \cite{video} and (c) TELCO \cite{telco}. Basic information about the datasets is summarized in Table \ref{tab:dataset}.

\begin{center}
    \begin{table}
    \centering
    \begin{tabular}{ | p{1.7cm} | p{2cm} | p{1.2cm} | p{1.2cm}|}
    \hline
    Dataset & Type & Number of Training Samples & Number of Test Samples \\ \hline
    Activity \cite{Samsung} & Classification  & 7352  &  2947\\ \hline
    Video \cite{video} & Regression & 45856 & 22928\\ \hline
    TELCO \cite{telco} &  & & \\  
    Periodic & Classification&25520 & 25523\\
    Flash & &7575 & 7575 \\
    Linear & &8252 & 8254 \\
    \hline
    \end{tabular}
    \caption{Datasets used in experiments}
    \label{tab:dataset}
    \end{table}
    \vspace{-0.5cm}
\end{center}

Samsung \cite{Samsung} is an activity recognition dataset with $6$ classes as labels. It consists of $560$ features. Video \cite{video} dataset has video transcoding time as the label and consists of $10$ features. TELCO \cite{telco} is a binary classification dataset with SLA violations as the label. It consists of $24$ features.

Activity \cite{Samsung} and Video \cite{video} datasets are used as is without normalization. Random forest classification and regression are used respectively for the Activity and Video datasets. These datasets are divided into train and test using random shuffling. The sources of these datasets do not claim to provide varying distributions. On the other hand, the TELCO \cite{telco} dataset inherently consists of different loads seen in the network. Ideally, this should be reflected in the data-deviation scores.

\begin{center}
    \begin{table}
    \centering
    \begin{tabular}{ | p{1.5cm} | p{1.5cm} | p{1.5cm} | p{1.5cm}| p{1.5cm} | p{1.5cm} | p{1.5cm} | p{1.5cm} |}
    \hline
    Number of Samples & RMSE & KL Divergence & Wasserstein Metric & Similarity & Confidence & Accuracy \\ \hline
    10 & 0.056  & inf & 0.03 & 1& 0.5&0.8\\ \hline
    20 & 0.033 & inf &0.018 & 1& 0.48&0.6\\ \hline
    30 & 0.051 & inf & 0.025& 0.89&0.5 &0.6\\ \hline
    40 & 0.029 & inf & 0.015& 1&0.5&0.72\\ \hline
    50 & 0.027 & inf & 0.013& 1&0.47 &0.7\\ \hline
    100 & 0.029 & inf&  0.014& 0.94&0.47&0.73\\ \hline
    200 & 0.018 & inf&  0.009& 0.98&0.49&0.78\\ \hline
    500 & 0.015 & inf &  0.008& 0.99&0.48&0.72\\ \hline
    1000 & 0.014 & 0.017& 0.008& 0.98 &0.49&0.74\\ \hline
    5000 & 0.013 & 0.017& 0.008&  0.98&0.48& 0.73\\ \hline
    1000 & 0.014 & 0.017& 0.008& 0.97&0.48 & 0.73\\ \hline
    2000 & 0.013 & 0.016& 0.007& 0.98&0.48& 0.73\\ \hline
    Correlation &-0.25 &- &-0.23 &0.41 & 0.05& \\
    \hline
    \end{tabular}
    \caption{Data deviation scores reported using the Samsung classification dataset for different number of samples. The correlation of these data deviation scores with algorithm accuracy is reported in the last row of the table. RMSE, KL Divergence and Wasserstein are expected to have a negative correlation (closer to -1 is better), while the similarity and confidence scores are expected to have a positive correlation (closer to +1 is better). It can be seen that {\it Similarity} score shows the highest correlation with predictive performance of the algorithm.}
    \label{tab:samsungSamples}
    \vspace{-5mm}
    \end{table}
\end{center}

\begin{center}
    \begin{table}
    \centering
    \begin{tabular}{ | p{1.5cm} | p{1.5cm} | p{1.5cm} | p{1.5cm}| p{1.5cm} | p{1.5cm} | p{1.5cm} | p{1.5cm} |}
    \hline
    Number of Samples & RMSE & KL Divergence & Wasserstein Metric & Similarity & $R^{2}$ \\ \hline
    10 & 0.118 & inf & 0.077& 0.77& 0.91\\ \hline
    20 & 0.109 & inf & 0.069& 0.78 &0.97\\ \hline
    30 & 0.095 & inf & 0.052& 0.66 &0.82\\ \hline
    40 & 0.066 & inf & 0.038& 0.69 &0.96\\ \hline
    50 & 0.081 & inf & 0.048& 0.69&0.96\\ \hline
    100 & 0.059 &inf & 0.034&0.69 &0.96\\ \hline
    200 & 0.049 & inf&  0.022&0.69&0.83\\ \hline
    500 & 0.036 & inf& 0.015&0.69 &0.9\\ \hline
    1000 & 0.04 & inf& 0.016&0.69 &0.89\\ \hline
    5000 & 0.033 & 0.085&  0.011&0.66 &0.9\\ \hline
    1000 & 0.032 & 0.089 & 0.011&0.65 &0.91\\ \hline
    2000 & 0.032 & 0.085 & 0.011&0.66 &0.91\\ \hline
    Correlation  & 0.17& - & 0.25& 0.4& \\
    \hline
    \end{tabular}
    \caption{Data Deviation scores reported using the video regression dataset. The correlation of these data deviation scores with algorithm accuracy is reported in the last row of the table. RMSE, KL Divergence and Wasserstein are expected to have a negative correlation (closer to -1 is better), while the similarity score is expected to have a positive correlation (closer to +1 is better). It can be seen that {\it Similarity} score shows the highest correlation with predictive performance of the algorithm.}
    \label{tab:videoSample}
    \vspace{-5mm}
    \end{table}
\end{center}
\vspace{-1cm}
\subsection{Experimental Setup}
We report the Similarity score along with other distribution divergence techniques such as RMSE, KL, and Wasserstein. For classification tasks, we also report the probability/confidence values that are inherently reported by most classification algorithms. We calculate the correlation that these scores have the with the predictive performance of the algorithms (accuracy for classification and $R^2$ for regression).

Note that a high value of RMSE, KL, and Wasserstein indicates a divergence in distribution, while a low value of Similarity and confidence indicates divergence. One expects the algorithm's predictive performance to drop when the inference distribution is different from training distribution. We report a correlation score between different metrics and the true accuracy of the algorithm across different datasets, with varying number of sample sizes. A negative correlation in case of RMSE, KL and Wasserstein (closer to -1) and a positive correlation in case of Similarity and Confidence (closer to 1) is desirable for practical usage of these metrics in production.

The deviation scores are calculated per each feature and there are several features in each dataset. 
%% DREW: this seems like an important point, maybe make it earlier as well?
However, the average score of only a few is chosen, based on the feature importance value reported by the algorithm/model being used to make the predictions. Feature importance is reported as a standard metric in most of the machine learning algorithms.

Agnostic of the number of samples, the data deviation score should indicate that there is no divergence between training and inference samples when using the datasets Samsung and Video as they belong to the same set. To validate this, we perform experiments with varying number of samples and report them in Table \ref{tab:samsungSamples} and \ref{tab:videoSample} for the Samsung and Video datasets respectively.

When the number of samples is low, it is observed that KL divergence struggles. This is due to the fact that some bins during the creation of inference histogram were empty causing the value to be infinite. Similarity score performs consistently well across both datasets for varying number of samples. Since all the samples belong to the same dataset, one would expect the data deviation scores to not vary much across different number of samples and this is the case with the {\it Similarity} score.

\begin{center}
    \begin{table*}
    \centering
    \begin{tabular}{ | p{1.7cm} |p{1.5cm} | l | p{1.5cm} | p{1.5cm}| p{1.5cm} | l | l | l |}
    \hline
    Train - Test Loads & Number of Samples & RMSE & KL Divergence & Wasserstein Metric & Similarity & Confidence & Accuracy \\ \hline
    \multirow{3}{4em}{Periodic-Flash} & 10 & 0.096 & inf & 0.038 & 1 & 0.76 & 1 \\ 
    & 50  & 0.065 & inf & 0.036 & 1 & 0.7 & 0.74\\ 
    & 100 & 0.057 & inf & 0.034 & 1 & 0.69 & 0.72\\ 
    & 200 & 0.056 & 0.2 & 0.034 & 1 & 0.72 & 0.76\\\hline
    \multirow{3}{4em}{Periodic-Linear} & 10 & 0.152 & inf & 0.077 & 1.0 & 0.65 & 0.5\\ 
    & 50 & 0.058 & inf & 0.025 & 1 &  0.69 & 0.32\\ 
    & 100  & 0.081 & 0.0399 & 0.037 & 1 & 0.7 & 0.36\\ 
    & 200 & 0.076 & 0.292 & 0.033 & 1 & 0.68 & 0.36\\\hline
    \multirow{3}{4em}{Flash-Periodic} & 10 & 0.071 & inf & 0.03 & 0.97 & 0.72 & 0.6\\
    & 50 &  0.06 & 0.242 & 0.009 & 0.89 & 0.67 & 0.38\\ 
    & 100 &  0.079 & 0.416 & 0.024 & 0.73 & 0.69 & 0.43\\ 
    & 200 & 0.068 & 0.214 & 0.024 & 0.71 & 0.7 & 0.4\\ \hline
    \multirow{3}{4em}{Flash-Linear} & 10 & 0.076 & inf & 0.03 & 1 & 0.78 & 0.9\\ 
    & 50 & 0.067 & inf & 0.015 & 1 & 0.75 & 0.82\\ 
    & 100 & 0.065 & inf & 0.023 & 1 & 0.75 & 0.8\\ 
    & 200 & 0.067 & inf & 0.024 & 1 & 0.74 & 0.84\\
    \hline
    \multirow{3}{4em}{Linear-Periodic} & 10 & 0.082 & inf & 0.039 & 0.69 & 0.86 & 0.34\\ 
    & 50 & 0.055 & 0.178 & 0.027 & 0.79 & 0.86 & 0.34 \\ 
    & 100 & 0.088 & 0.273 & 0.038 & 0.62 & 0.86 & 0.31 \\ 
     & 200 & 0.1 & 0.332 & 0.041 & 0.59 & 0.86 & 0.35 \\ \hline
    \multirow{3}{4em}{Linear-Flash} & 10 & 0.067 & inf & 0.028 & 1 & 0.87 & 0.6\\ 
    & 50 & 0.083 & inf & 0.029 & 0.74 & 0.86 & 0.58\\
    & 100 & 0.069 & inf & 0.024 & 0.77 & 0.85 & 0.5\\
    & 200 & 0.059 & 0.236 & 0.027 & 0.79 & 0.86 & 0.54\\ \hline
    Correlation &  & -0.09 & - & -0.07 & 0.58 & -0.1&\\
    \hline
    \end{tabular}
    \caption{Data deviation scores reported using the TELCO dataset for different number of samples across different training and test loads. The correlation of these data deviation scores with algorithm accuracy is reported in the last row of the table. RMSE, KL Divergence and Wasserstein are expected to have a negative correlation (closer to -1 is better), while the similarity and confidence scores are expected to have a positive correlation (closer to +1 is better). It can be seen that {\it Similarity} score shows the highest correlation with predictive performance of the algorithm.}
    \label{tab:telco_samplenum}
    \vspace{-4mm}
    \end{table*}
    \vspace{-0.1in}
\end{center}

We also perform the same experiment with the TELCO dataset across different loads and the results are reported in Table
%% DREw: table names not showing up
\ref{tab:telco_samplenum}. Again as expected, the Similarity score shows much better correlation compared to other methods. 

In order to evaluate how the different scores behave in the presence of data corruption, we add a varying amount of random noise (which is equal to one standard deviation of each feature respectively) to the datasets. We report the results in Table \ref{tab:samsungNoise} and \ref{tab:videoNoise} respectively. All the metrics including the Similarity score perform equally well.

\begin{center}
    \begin{table}
    \centering
    \begin{tabular}{ | p{1cm} | l | l | l| l | l | l | l |}
    \hline
    Noise Level & RMSE & KL & Wasserstein Metric & Similarity & Confidence & Accuracy \\ \hline
    0 & 0.011 & 0.015 &0.006 & 0.99& 0.48&0.72\\ \hline
    0.1 & 0.018 & 0.034 & 0.01& 0.92& 0.47& 0.72\\ \hline
    0.2 & 0.031 & 0.067 & 0.017& 0.85&0.46 &0.69\\ \hline
    0.3 & 0.039 & 0.1 &  0.021& 0.8&0.45 &0.68\\ \hline
    0.4 & 0.049 & 0.144 & 0.027& 0.75&0.43&0.66\\ \hline
    0.5 & 0.062 &0.202 & 0.034& 0.67& 0.42&0.65\\ \hline
    0.6 & 0.071 &0.262 & 0.039& 0.62& 0.41&0.63\\ \hline
    0.7 & 0.083 & 0.335& 0.045& 0.55& 0.39&0.61\\ \hline
    0.8 & 0.094 & 0.42&  0.051& 0.5&0.38 &0.6\\ \hline
    0.9 & 0.104 & 0.514& 0.055& 0.44&0.37&0.56\\ \hline
    Correlation & -0.99 & -0.98& -0.98& 0.98& 0.98& \\
    \hline
    \end{tabular}
    \caption{Data deviation scores reported using the Samsung classification dataset when noise is added. Data deviation scores reported using the TELCO dataset for different number of samples across different training and test loads. The correlation of these data deviation scores with algorithm accuracy is reported in the last row of the table. RMSE, KL Divergence and Wasserstein are expected to have a negative correlation (closer to -1 is better), while the similarity and confidence scores are expected to have a positive correlation (closer to +1 is better). It can be seen that all the metrics perform equally well.}
    \label{tab:samsungNoise}
    
    \end{table}
\end{center}
\begin{center}
    \begin{table}
    \centering
    \begin{tabular}{ | l | l | l | l| l | l | l | l |}
    \hline
    Noise Level & RMSE & KL  & Wasserstein Metric & Similarity & $R^{2}$ \\ \hline
    0 & 0.006 & 0.023 & 0.004 & 1 & 0.85 \\ \hline
    0.1 & 0.005 & 0.016 & 0.003 & 1 & 0.71 \\ \hline
    0.2 & 0.007 & 0.02 & 0.003 & 0.98& 0.57 \\ \hline
    0.3 & 0.009 & 0.028 & 0.005& 0.97 & 0.42 \\ \hline
    0.4 & 0.013 & 0.04 & 0.007& 0.95& 0.31\\ \hline
    0.5 & 0.017 & 0.052 & 0.009&0.94 & 0.14 \\ \hline
    0.6 & 0.019 & 0.067 & 0.01& 0.92& 0.01\\ \hline
    0.7 & 0.022 & 0.086 & 0.012& 0.91& -0.1 \\ \hline
    0.8 & 0.026 & 0.11 & 0.014& 0.89& -0.31\\ \hline
    0.9 & 0.031 & 0.143 & 0.016& 0.87& -0.47\\ \hline
    Correlation  & -0.98 & -0.95 & -0.97 & 0.99 & \\
    \hline
    \end{tabular}
    \caption{Data deviation scores reported using the video regression dataset. Data deviation scores reported using the TELCO dataset for different number of samples across different training and test loads. The correlation of these data deviation scores with algorithm accuracy is reported in the last row of the table. RMSE, KL Divergence and Wasserstein are expected to have a negative correlation (closer to -1 is better), while similarity is expected to have a positive correlation (closer to +1 is better). It can be seen that all the metrics perform equally well.}
    \label{tab:videoNoise}
    \vspace{-4mm}
    \end{table}
\end{center}
\vspace{-1cm}
\subsection{Results}
\label{subsec:results}
Experiments performed across three datasets demonstrate the following key takeaways:

\begin{enumerate}
    \item Similarity score performs consistently well in the presence and absence of noise agnostic of the number of samples present in inference. This is demonstrated by the results across three datasets, briefly described below.
    \item Similarity score shows highest correlation with the predictive performance of the algorithm/model when a small number of samples are present in the incoming data across the three datasets. This can be
    %% DREW: fill in table
    seen from results presented in Table \ref{tab:samsungSamples} and \ref{tab:videoSample}. Note that the incoming data belongs to the same dataset and was not modified in any way.
    \item Similarity score shows a high correlation with the predictive performance of the algorithm/model when the nature of incoming data changes and the number of samples is small. This is demonstrated using the TELCO dataset across three different loads. This can be
    %% DREw: fill in table
    seen from results presented in Table \ref{tab:telco_samplenum}.
    \item All the data deviation scores perform equally well when there is data corruption, irrespective of the number of samples in the incoming data across all three datasets. This can be seen from results presented in Table \ref{tab:samsungNoise} and \ref{tab:videoNoise}. Noise was explicitly added to these datasets.
\end{enumerate}

A key difference between other scores reported in the experiments (RMSE, KL, and Wasserstein) and the Similarity score is that the other metrics were designed to quantify the distance between two distributions. We presented an alternate scoring technique {\it Similarity} that does not rely on inference distribution and addresses the issues pointed out in Section \ref{sec:health}.

\if 0
This has two drawbacks (a) To achieve a high score, inference distribution as a whole needs to be close to training. This is problematic as it is possible that only a few types of patterns seen during training might appear during inference. Such a case should not be penalized. Training datasets are created with the goal to cover all possible foreseeable patterns, of which only a subset might occur at a time during production deployment. (b) When there are only a few samples available at a time as is the case with streaming and some batch applications where the batch size varies, a representative histogram is hard to create. Similarity score mitigates both the issues by using the probability score of the multinoulli distribution directly. This makes the score (a) penalize divergences asymmetrically where high occurrence of under-represented categories is penalized while the high occurrence of over-represented categories is not, within each feature (b) The score works well across different sample size ranges as it does not rely on representative inference histogram creation. 
\fi
\subsection{System Deployment and Operations}
\label{subsec:sysDeploy}
We now describe how a data scientist deploys his/her existing code/pipeline via the MCenter system. 

As described earlier, data scientists can bring any existing code to be deployed by MCenter. The data scientist should include a JSON file (i.e., file containing a description of the input/output arguments, engine/library it is built on, and any additional dependencies) along with the code/pipeline that needs to be deployed. The JSON file along with the code and dependent libraries need to be packaged into a compressed file (such a zip or tar) and uploaded into the system.
Additionally, one can add instrumentation to the code to output metrics that can be tracked when the code is executed via MCenter. To simplify the effort, MCenter comes with language-agnostic MLOps API.

The MLOps API helps users to report and fetch health and other user-defined metrics. In the context of this paper, distributions of input data of both training and inference can be reported with a single call that takes a dataframe (or an equivalent object) as input. The system comes with a built-in Similarity score calculator and also enables users to provide their own custom code to calculate the similarity score. When the code is executed via the MCenter system, the statistics (such as data distribution and similarity scores) are automatically reported. These statistics can also be programmatically fetched via a separate pipeline. All of the reported metrics and statistics are automatically displayed in a web-based dashboard. For example, the overlapping data distribution histograms between training and inference pipelines along with their Similarity score are automatically displayed in the dashboard. 

Thresholds can be set to identify and report data deviation between training and inference pipelines. A data scientist has the ability to set thresholds on the individual (or a group of) attributes of the incoming data. These thresholds can be used to generate health alerts. It is important to understand that in any practical deployment, it's challenging to know how to set these threshold values initially. A data scientist can always update the threshold values based on Similarity scores after observing the system over a period of time. Note that for datasets with hundreds of attributes, is it not always possible to manually set thresholds. To mitigate these issues, MCenter provides an option to auto-configure the threshold values to be $+ \epsilon$ of the similarity values produced by the first few inference runs. 

\section{Conclusions and Future Work}
We presented the framework of ML Health and explain why addressing it is necessary for production ML workflows. Within this framework, we presented an ML Health metric, {\it Similarity},  as a first step towards monitoring the fitness of ML models deployed in production. Using the Similarity score, we demonstrated desirable characteristics, namely its asymmetry and its ability to effectively detect data deviations with a small number of samples. Our results demonstrate how Similarity outperforms known state of the art comparative metrics.

%% DREW: Re-emphasize how this does better than other algorithms
Providing a system architecture that enables the propagation of such metrics is key for adoption of such techniques in production monitoring. To inter-operate with the diverse developments in ML/DL, we created a system that integrates with several popular engines and languages. This system is already in use by enterprise products and uses the Similarity score metric to monitor the fitness of ML models in production to inference data.

We currently limit ourselves to the univariate analysis as a baseline approach and will expand to a multivariate approach to capture the variation in relationship between different features for both categorical and continuous types. Datasets used for natural language processing (NLP) are extremely sparse and need special techniques to detect divergence in feature patterns. Similarly, image datasets that typically use deep learning algorithms also need techniques that can detect a divergence in incoming patterns. These are interesting directions that we would like to pursue in the future. 

Monitoring model performance in the absence of labels is an important problem being faced in production with the increasing adoption of machine learning techniques. 
%% DREw: this sentence comes out of nowhere. Which issue? monitoring?
%% What other techniques? examples?
We believe that there is a tremendous amount of scope for research in this direction. Rather than relying on a single metric, it would be beneficial to develop an ensemble of techniques that track model performance for different domains such as NLP, images, compression etc. Existence of several techniques enables combining and customization for different applications.

%\end{document}  % This is where a 'short' article might terminate

%
% The next two lines define the bibliography style to be used, and the bibliography file
\bibliographystyle{unsrt}  
\bibliography{references}  %%% Remove comment to use the external .bib file (using bibtex).
%%% and comment out the ``thebibliography'' section.

\newpage
% 
% If your work has an appendix, this is the place to put it.
\appendix

\section{Reproducibility}
All the results presented in the paper are based on publicly available datasets. Code to generate them is in a public git repository and steps to reproduce them will be described in this Section. 

The MCenter system that uses this score in a production deployment is also available for free download. It comes in-built with sample components in Python, PySpark, R and TensorFlow. Users can upload their own code with custom statistics and visualize them in production using the MCenter System. We also provide a brief description of the MCenter system in this Section. The documentation that comes with the free version of the system contains details of all the features/aspects of MCenter, which are out of the scope of this paper.

\subsection{Code}
All the code for reproducing the experimental results reported in this paper was developed in Python and should be executed using Python3. Code described in the following paragraphs is available at : \\ \hyperref[]{https://github.com/mlpiper/mlhub/tree/kdd2019}.

First step towards reproducing the experiments is the creation of training and test datasets from the raw files available at the public repositories. We provide $3$ python scripts, one per each dataset: (a) Samsung \cite{Samsung} (b) Video \cite{video} (c) telco \cite{telco} can use (a) samsung\_dataset\_creation.py (b) video\_dataset\_creation.py (c) telco\_ds\_double\_algo\_exp.py respectively available the ml\_health folder of the repository. Note that the paths to data-files within these files will need to be provided depending on where the user downloads the raw datasets in the local system. 

Once the training and test files have been created for each dataset, all the results reported in the Tables of the paper can be reproduced using the Python script: {\it KDD2019.py}. Note that the file paths will again need to be provided depending on where the user stores these datasets in the local system.

\subsection{System Access}
In this section, we describe how the ml health of the system can be used by a data scientist for deployment in production. Free access of MCenter is avaialble at:\\
\hyperref[]{''https://www.parallelm.com/free-account/''}. This spins up an Azure system pre-configured to support Spark, TensorFlow, PySpark, Scikit-Learn, and R. It can deploy built-in containers and also deploy user-defined containers (from their container registry). The login screen of the system is shown in Figure \ref{fig:mcenter}. Each part of the system is briefly described below:

\begin{figure}[htp!]
\centering
\includegraphics[trim={0cm 0cm 0cm 0cm},clip,width=0.8\textwidth]{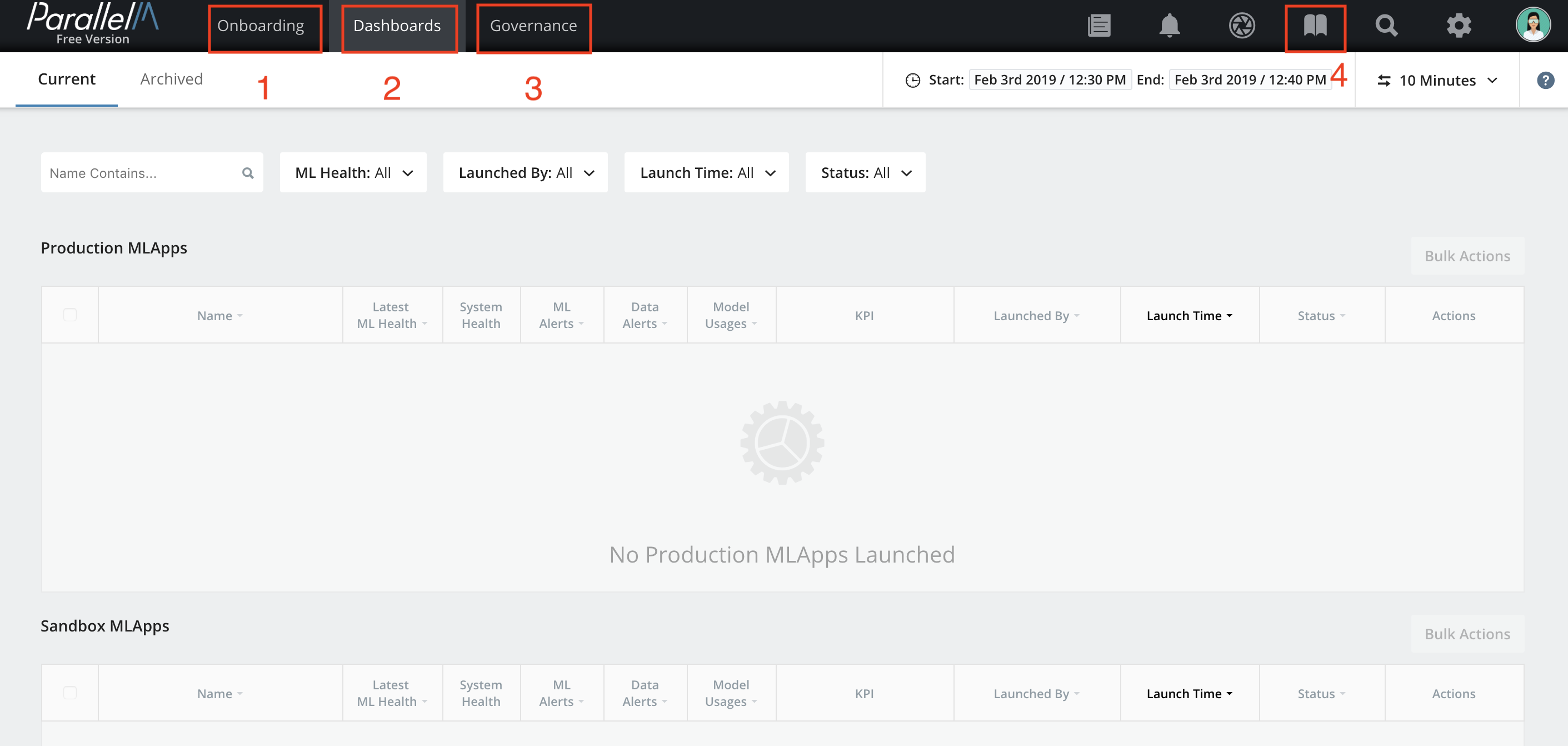}           
\caption{Login Screen of MCenter \label{fig:mcenter}}
\label{fig:secon}
\end{figure}

\begin{enumerate}
\item Deployment: This tab shows all the MLApps running in the system in a central location.
\item Dashboards: This tab consists of components, pipelines, pattern and profiles that exist in the MCenter system.
\item Governance: This tab shows all the models produced/uploaded in this system along with the details of their usage and relationship to the pipelines.
\item Documentation: This section describes all the features offered by MCenter to make production deployment of Machine Learning Algorithm smooth for both operations and data science teams.
\end{enumerate}

\begin{figure}[htp!]
\centering
\includegraphics[trim={0cm 0cm 0cm 0cm},clip,width=0.8\textwidth]{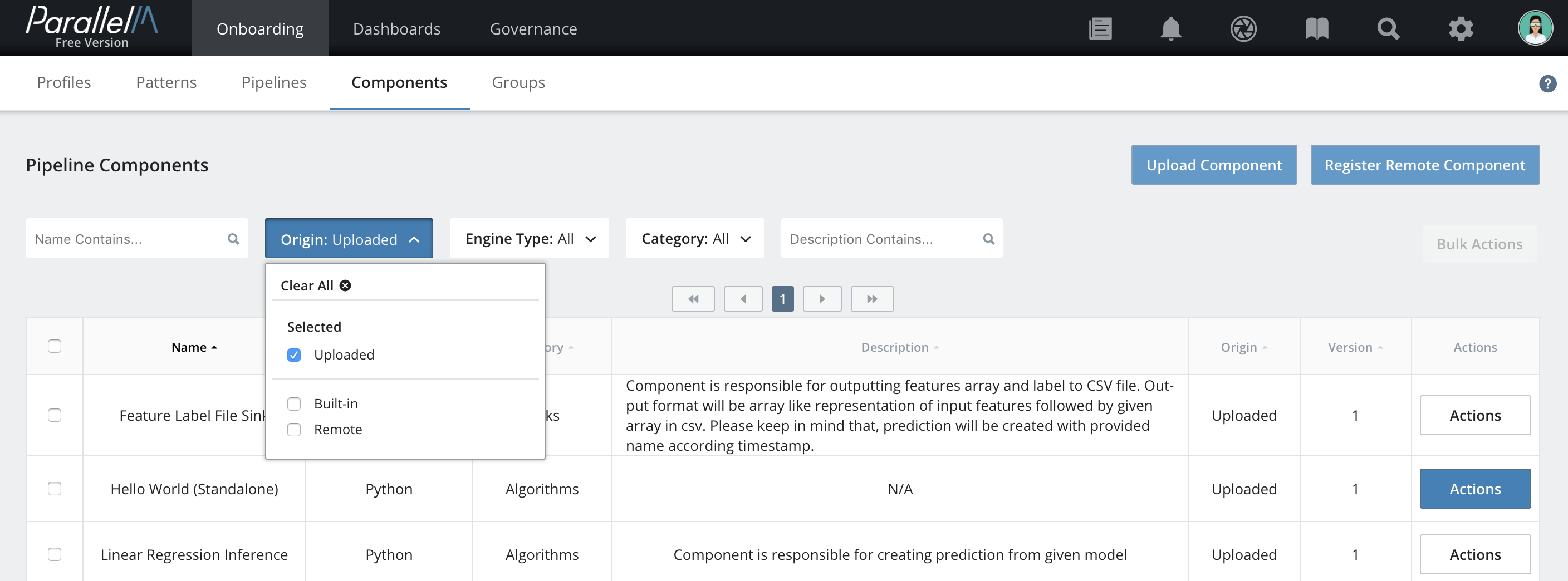} 
\caption{Components in MCenter} \label{fig:deployment}
\end{figure}

We will focus on the Deployment aspect of the product which enables a data scientist to bring their code into the system. Figure \ref{fig:deployment}, shows various tabs within Deployment, whose detailed description can be found in the product documentation. One can view the code in sample components that come pre-uploaded and create/upload custom components. Each part of the system is briefly described below:

\begin{enumerate}
\item Upload Component: One can upload a custom component using this option.
\item Upload Model: An option to upload a externally trained model is avaialble in the Governance tab similar to the upload component option in the components tab.
\item Actions: One can take actions such as downloading/viewing or deleting a component using the actions button available with each component.
\end{enumerate}

Figure \ref{fig:profile} describes the construction of an MLApp within the system for deployment. Each part of the MLApp is briefly described below:

\begin{figure}[htp!]
\centering
\includegraphics[trim={0cm 0cm 0cm 0cm},clip,width=0.4\textwidth]{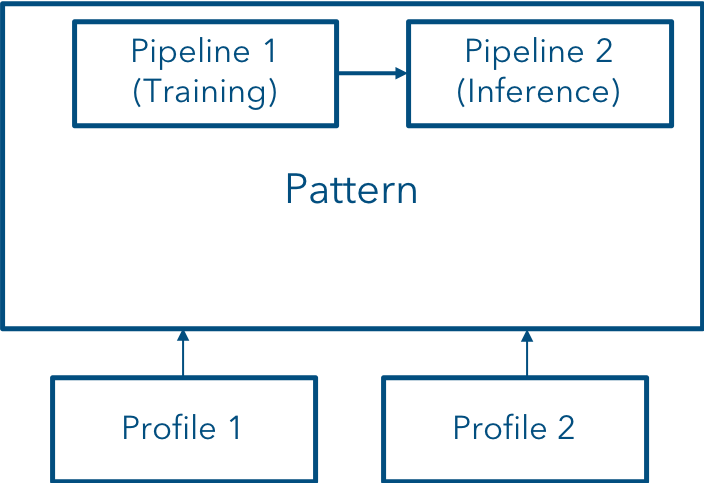}           
\caption{Components of an MLApp \label{fig:profile}}
\end{figure}

\begin{enumerate}
    \item Component: It is the basic unit of computation. It contains a piece of code that can run on your  favorite engine such as TensorFlow, Spark, etc.
    \item Pipeline: It facilitate re-using common components. It contains either a standalone or connected components with additional parameters such as schedules to run.
    \item Pattern: It relates pipelines into a logical unit. A Pattern can contain one or more pipelines that may transfer models between them. Additional information such as transfer policies can be specified in a pattern.
    \item Profile: It allows customization of patterns which enable scale. A profile is used to create instances of a pattern with different parameters. Example: Profile 1 could train with a different regularization parameter compared to Profile 2.  

\end{enumerate}

A profile can be launched either in production or sandbox after configuring parameters of the pipeline and individual schedule.

\end{document}